\pdfoutput=1

\documentclass[11pt]{article}

\usepackage[]{ACL2023}

\usepackage{times}
\usepackage{latexsym}

\usepackage[T1]{fontenc}
\usepackage{tipa}
\newcommand{\ny}{\textltailn}  

\usepackage[utf8]{inputenc}

\usepackage{microtype}

\usepackage{inconsolata}

\usepackage{graphicx}

\usepackage{tabularx}
\usepackage{multirow}
\usepackage{eso-pic}

\usepackage{amsmath}
\usepackage{tcolorbox}
\title{Feriji: A French-Zarma Parallel Corpus, Glossary \& Translator}

\author{
Mamadou K. KEITA\textsuperscript{},
   Elysabhete Amadou Ibrahim\textsuperscript{1}, Habibatou Abdoulaye Alfari\textsuperscript{1},
   Christopher Homan\textsuperscript{2}\\
  \textsuperscript{1}Ashesi University\\
  \textsuperscript{2}Rochester Institute of Technology\\
}

\begin{document}
\maketitle

\begin{abstract}
Machine translation (MT) is a rapidly expanding field that has experienced significant advancements in recent years with the development of models capable of translating multiple languages with remarkable accuracy. However, the representation of African languages in this field still needs improvement due to linguistic complexities and limited resources. This applies to the Zarma language, a dialect of Songhay (of the Nilo-Saharan language family) spoken by over 5 million people across Niger and neighboring countries \cite{lewis2016ethnologue}. This paper introduces Feriji, the first robust French-Zarma parallel corpus and glossary designed for MT. The corpus, containing 61,085 sentences in Zarma and 42,789 in French, and a glossary of 4,062 words represents a significant step in addressing the need for more resources for Zarma. We fine-tune three large language models on our dataset, obtaining a BLEU score of 30.06 on the best-performing model. We further evaluate the models on human judgments of fluency, comprehension, and readability and the importance and impact of the corpus and models. Our contributions help to bridge a significant language gap and promote an essential and overlooked indigenous African language.

\end{abstract}

\section{Introduction}
The field of MT has witnessed substantial progress, particularly with the development of sophisticated models capable of accurately translating multiple languages. These models sometimes even get closer to human proficiency \cite{Farahani2020AdequacyIM}. However, despite these advances, African languages still need representation in MT systems, primarily due to linguistic complexities and limited resources \cite{joshi2020state}.
One such under-represented language is Zarma, spoken by over 5 million people, predominantly in Niger \cite{ethnologue2023}. As a member of the Songhay family within the Nilo-Saharan language group, Zarma has received limited attention in natural language processing research. This lack of representation restricts Zarma speakers' access to technology and hinders efforts to preserve and promote Zarma.
To address this challenge, we introduce Feriji---the first parallel French-Zarma corpus and glossary designed specifically for MT tasks. The corpus contains 61,085 sentences in Zarma and 42,789 in French, representing a significant step towards enriching MT resources for the Zarma language. The development of Feriji involved extensive collection, alignment, and cleaning of texts, resulting in a resource that not only bridges a significant linguistic gap but also promotes the use of Zarma in research contexts. We chose French as the source language because Niger is a French-speaking country, and most information and resources are readily available in French rather than any other language. This makes French a practical choice for creating a resource that can effectively support the Zarma-speaking community.
This paper details the creation process of Feriji, structure, and potential value for MT research, particularly for the Zarma language. By providing this resource, we aim to facilitate further research in this area and enhance the integration of Zarma into the global MT field.

\section{Literature Review}
Advances in MT have been a significant focus within natural language processing (NLP). In recent years, we have seen the rise of neural machine translation (NMT) models capable of producing translations that approach---or even surpass---human proficiency in many languages. Models such as Facebook's M2M-100 \cite{fan2020beyond, schwenk2019ccmatrix, el2019massive} have revolutionized multilingual translation with their accuracy. However, the representation of African languages in MT remains a significant challenge, as highlighted in several studies \cite{10.1145/3567592}.

African languages, numbering approximately 3,000, are diverse and complex, characterized by unique tonal nuances and dialects \cite{lewis2016ethnologue}. Representing these languages in MT systems is a substantial task, requiring extensive resources and expert input.
The under-representation of African languages in MT systems is particularly concerning, given the literacy rates in Sub-Saharan Africa. As of 2020, the literacy rate stood at 67.27\%, while in Niger, it is at 80.9\% as of 2023 \cite{worldbank2023literacy}. This data indicates that a significant portion of the population relies on native languages for communication, unlike in regions with higher literacy rates. The comparatively high illiteracy rates further highlight the importance of including these native languages in initiatives through translation systems.

Efforts to address the under-representation of African languages include initiatives like the Masakhane project, which focuses on strengthening NMT for African languages \cite{nekoto2020participatory}; the Aya Model \cite{Ustun2024AyaMA}, a multi-task model covering 101 languages (over 50\% of which are low-resource); and Facebook's No Language Left Behind (NLLB) project \cite{nllb2022}, which aims to enable translation into over 60 African languages. Unfortunately, no specific initiative has targeted Zarma or any Songhay language, leaving them largely unexplored in the MT field.

This literature review highlights the importance of our work in contributing to the diversification of language resources in MT, particularly for low-resource languages such as Zarma.

\section{Feriji}

\subsection{Feriji Dataset}
The Feriji Dataset (FD)\footnote{\url{https://github.com/27-GROUP/Feriji/tree/main/feriji/zar_fr_sentences}} is a parallel corpus of French and Zarma sentences designed for machine translation tasks. The dataset currently contains 42,789 French sentences and 61,085 Zarma sentences, all grouped into aligned entries---each entry consists of sentences in one language paired with its corresponding translation in another. The dataset is split into training, validation, and test sets with an 80/10/10 split.
Linguistically, the dataset comprises 794,709 words in French and 847,362 words in Zarma. The French portion exhibits higher lexical diversity, with 21,592 unique words compared to 9,902 unique words in the Zarma portion. This vocabulary size difference reflects the two languages' varying linguistic richness within the dataset. Additional insights into the dataset's characteristics are presented in Tables \ref{tab:dataset_glossary_counts} and \ref{tab:Sentence_length}.

\subsection{Feriji Glossary}
The Feriji Glossary (FG)\footnote{\url{https://github.com/27-GROUP/Feriji/tree/main/feriji/zar_fr_glossary}} is an important component of Feriji, containing 4,062 words. The glossary was curated to support the translation process between French and Zarma. This provides a valuable resource for both language learners and MT developers. The glossary entries were sourced primarily from extensive online resources, including the Bible, and supplemented by translations contributed by our team. This comprehensive collection of words and expressions not only aids in the translation process but also acts as a bridge between the two languages, enhancing understanding and communication between French and Zarma speakers. Including the glossary within Feriji significantly enriches its utility and robustness. This makes it a valuable resource for MT research and linguistic studies involving these two languages.

\begin{table}[!htb]
\centering
\small
\begin{tabular}{ c c c }
\hline
 & \textbf{French} & \textbf{Zarma}\\
\hline
\textit{Sentence Count} & 42,789& 61,085 \\
\textit{Glossary Word Count} & 4,062 & 4,062 \\
\textit{Number of Unique Words in FD} & 21,592 & 9,902 \\
\hline
\end{tabular}
\caption{Feriji Dataset and Glossary Statistics}
\label{tab:dataset_glossary_counts}
\end{table}
 
\begin{table}[!htb]
\centering
\small
\begin{tabular}{ c c c c }
\hline
 & \textbf{Word Range} & \textbf{French} & \textbf{Zarma}\\
\hline
Short Sentence & 1-5 words & 4,133 & 9,291 \\
Medium Sentence & 6-10 words & 8,048 & 15,388 \\
Long Sentence & 11+ words & 30,608 & 36,406 \\
\hline
\end{tabular}
\caption{Sentence Length Distribution in Feriji Dataset}
\label{tab:Sentence_length}
\end{table}

\subsection{Data Collection Pipeline}
The creation of FD involved a comprehensive sentence collection process from various sources. The primary sources included religious texts,\footnote{\url{http://visionneuse.free.fr}} materials from the Peace Corps,\footnote{\url{http://www.bisharat.net/Zarma/ZEF-L.htm}} and original stories generated using ChatGPT4 \cite{openai2023gpt4}, which were then translated by our team. The initial data contained noise and missing translations, which hindered its effectiveness. We employed a series of data cleaning and alignment scripts to address these challenges.

\begin{figure}[ht]
    \centering
    \includegraphics[width=8cm]{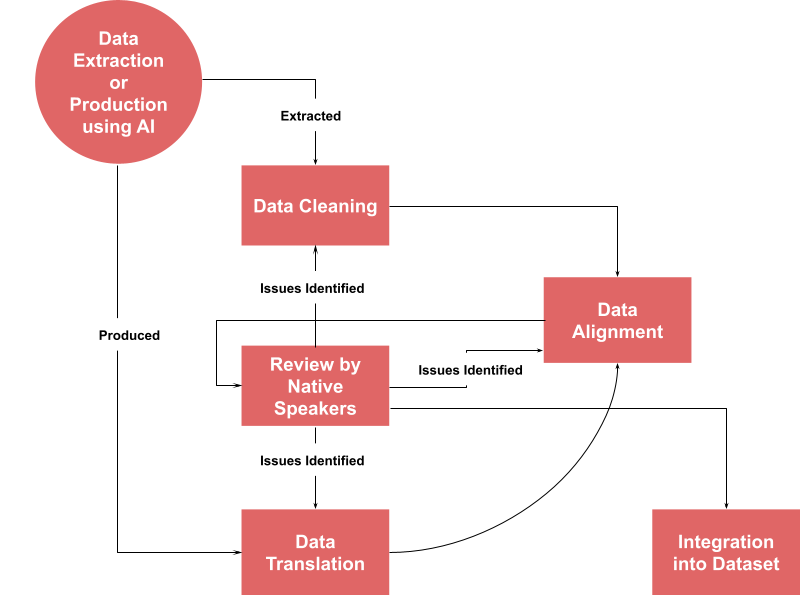}
    \caption{Data Collection Process}
    \label{fig:pipeline}
\end{figure}

After the initial alignment process, we conducted a human review phase in which Zarma speakers reviewed the aligned sentences to verify their accuracy. This process, as illustrated in Figure \ref{fig:pipeline}, ensures the viability of FD as a resource for both linguistic study and translation tasks. More details about the data collection, distribution across sources, and evaluation is explained in Section \ref{Detailed data collection}.

\section{Feriji-based Machine Translation}
To evaluate the effectiveness of Feriji, we fine-tuned three state-of-the-art language models on the French-to-Zarma translation task: MT5-small \cite{xue2020mt5}, M2M100, and NLLB-200-distilled-600M (NLLB-200-dist). We used a P100 GPU in the Kaggle environment for training. Table \ref{tab:training_results} presents the results of our experiments.

\subsection{Model Selection and Parameters}
The candidate models for fine-tuning were selected based on their multilingual capabilities, which are crucial for handling the complexities of Zarma translation and the ease of training them on our dataset. Below is a brief overview of the models, their parameters, and the rationale behind their selection.

\subsubsection{MT5-Small}
The MT5-Small model, a variant of the original T5 model, is specifically designed for multilingual tasks. With 300 million parameters, MT5-Small is equipped to handle various language translation tasks effectively. We chose MT5-Small because of its ability to accurately process and translate multiple languages.

\subsubsection{M2M100}
The M2M100 model stands out for its ability to translate directly between multiple languages without relying on English as an intermediary. Its 418 million parameters make it a robust model capable of handling the complexities of multilingual translation. M2M100's extensive language coverage makes it a suitable candidate for translating Zarma, as it can leverage learned patterns from other languages.

\subsubsection{NLLB-200-dist}
The NLLB-200-dist model is a distilled---and therefore more computationally efficient---version of the NLLB 600M model. With 600 million parameters in its original form, this model is expected to capture the nuances essential for accurately translating low-resource languages like Zarma better than smaller models. Its capacity to process various languages, including those with limited resources, aligns well with the goals of our project.

\begin{table}[h]
\centering
\begin{tabular}{ c c c c }
\hline
\textbf{Model}    & \textbf{Epoch} & \textbf{BLEU} \\
\hline
\textit{MT5-small}      & 20             &  6.10      \\
\textit{M2M100}         & 4              & 30.06         \\
\textit{NLLB-200-dist}  & 8              & 29.68       \\
\hline
\end{tabular}
\caption{Training Epoch and Results Across Models}
\label{tab:training_results}
\end{table}

The fine-tuning experiments yielded encouraging results, particularly for an early version of the Translator. The mean BLEU \cite{10.3115/1073083.1073135} score was 21.95, with the M2M100 model achieving the highest score of 30.06. These results demonstrate the effectiveness of FD and highlight potential areas for future improvement. Table \ref{tab:model_translation_comparison} provides example translations generated by the different models.

A major concern is the significantly lower performance of the MT5-small model compared to the M2M100 and NLLB-200-dist models. The primary reasons for this discrepancy are the smaller parameter size and less sophisticated pre-training data of the MT5-small model. With only 300 million parameters, MT5-small may not capture the intricate linguistic nuances required for accurate Zarma translations as effectively as the larger models with 418 million (M2M100) and 600 million (NLLB-200-dist) parameters.

Another aspect worth noting is the choice of hyperparameters. We used a consistent set of hyperparameters across all models to maintain fairness in comparison. However, it is possible that the MT5-small model may require a more adapted hyperparameter tuning process to optimize its performance for Zarma translation tasks.

\begin{table}[h]
\centering
\small
\begin{tabularx}{\columnwidth}{ X|X|X|X }
\hline
\textbf{Sentence} & \textit{MT5-small} & \textit{M2M100} & \textit{NLLB-200-distilled-600M} \\
\hline
Je suis devant la porte & Ay go fu meyo jine & Ay go fuo jine & Ay go meyo jine \\
Adeem et Habi partent à la maison & da Adem da Habi koy fuwo do &  Adem da Habi ga koy fu & Adeem da Habi ga koy fu \\
\hline
\end{tabularx}
\caption{Translation Comparison Across Models}
\label{tab:model_translation_comparison}
\end{table}


\section{Human Evaluation}
Since the BLEU metric alone cannot fully assess performance in our case, we conducted a human evaluation experiment to assess the quality of the translations produced by the NLLB-200-dist and M2M100 models. We recruited five native Zarma speakers to participate in the evaluation. Each participant received a set of 100 sentences that both models had translated. Participants rated each translation on a scale of 1 to 5 for fluency, accuracy, and readability, with 5 being the highest score:
\begin{itemize}
    \item \textbf{Fluency:} Assessed how natural and grammatically correct the translation sounded.
    \item \textbf{Accuracy:} Measured how accurately the translation conveyed the meaning of the original sentence.
    \item \textbf{Readability:} Evaluated how easy it was to read and understand the translation.
    \item The results of the human evaluation are presented in Tables \ref{tab:human_evaluation} and \ref{table:detailed_evaluation}. The M2M100 model produced translations rated as significantly more fluent, comprehensible, and readable than the translations produced by the NLLB-200-dist model.
\end{itemize}

\begin{table}[ht]
\centering
\scriptsize
\renewcommand{\arraystretch}{2}
\begin{tabular}{ c p{1cm} p{1.3cm} p{1.2cm} c }
\hline
\textbf{Model} & \textbf{Fluency} & \textbf{Accuracy} & \textbf{Readability} & \textbf{Total} \\
\hline
\textit{M2M100} & 4.2 & 4.1 & 4.0 & 12.3 \\
\textit{NLLB-200} & 3.5 & 3.6 & 3.4 & 10.5 \\
\hline
\end{tabular}
\caption{Human Evaluation Scores}
\label{tab:human_evaluation}
\end{table}

\begin{table*}[t]
\centering
\small
\begin{tabular}{c|c|c|c|c|c|c|c|c }
\hline
\textbf{Model} & \textbf{Metric} & \multicolumn{5}{c|}{\textbf{Annotator Scores}} & \textbf{var} & \textbf{Std. Dev.} \\ \cline{3-7}
& & A1 & A2 & A3 & A4 & A5 & & \\ \hline
\textbf{M2M100} & Fluency & 4 & 4 & 4 & 5 & 4 & 0.2 & 0.45 \\
& Comprehension & 4 & 4 & 4 & 4 & 5 & 0.2 & 0.45 \\
& Readability & 4 & 4 & 4 & 4 & 4 & 00 & 0.00 \\ \hline
\textbf{NLLB-200} & Fluency & 3 & 4 & 3 & 4 & 3 & 0.3 & 0.55 \\
& Comprehension & 3 & 4 & 3 & 4 & 4 & 0.3 & 0.55 \\
& Readability & 3 & 3 & 4 & 3 & 4 & 0.3 & 0.55 \\ \hline
\end{tabular}
\caption{Individual Score Details}
\label{table:detailed_evaluation}
\end{table*}

These findings suggest that the M2M100 model can better capture the nuances of the Zarma language and produce more faithful and readable translations of the original French text.

\section{Feriji Translator}
\begin{figure}[ht]
    \centering
    \includegraphics[width=8cm, height=3.2cm]{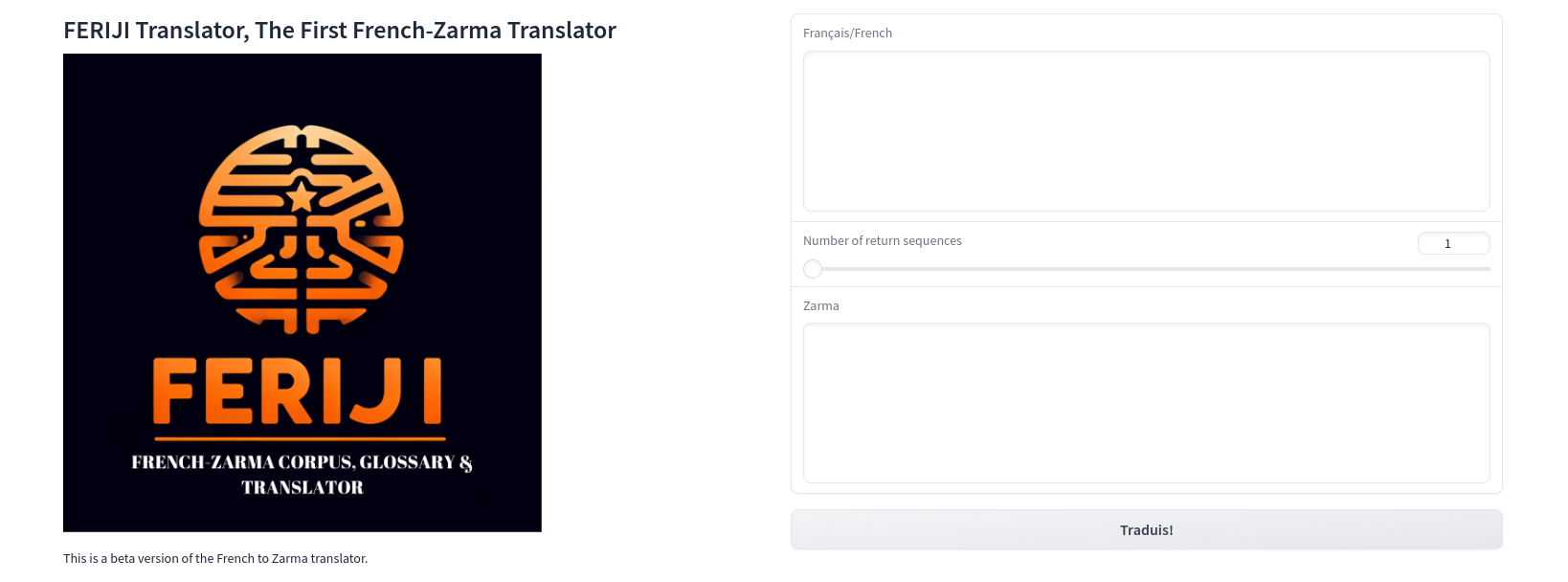}
    \caption{Feriji Translator Beta Interface}
    \label{fig:FT}
\end{figure}

The Feriji Translator (FT)---French to Zarma translator---is a crucial component of the Feriji project. It provides a means for non-native speakers to explore the Zarma language and for Zarma speakers to access textual resources available in French but not in Zarma. We chose the M2M100 model for FT because it achieved the highest BLEU score and performed well in our human evaluation, as shown in Tables \ref{tab:training_results} and \ref{tab:human_evaluation}. Figure \ref{fig:FT} shows the interface of the FT.

\section{Community Engagement and Feedback}
\begin{figure}[ht]
    \centering
    \includegraphics[width=8cm]{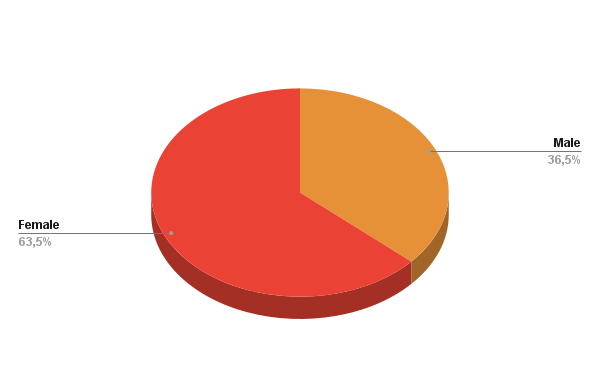}
    \caption{Gender representation in the survey}
    \label{fig:gender}
\end{figure}

\begin{figure}[ht]
    \centering
    \includegraphics[width=8cm]{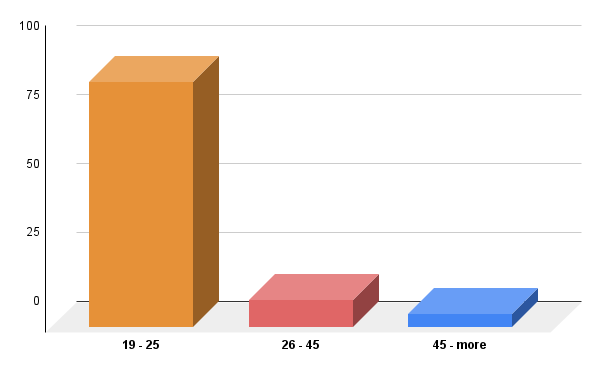}
    \caption{Age representation in the survey}
    \label{fig:age}
\end{figure}

Following the release of the FT and its associated model pipeline, we surveyed to gather feedback from the Zarma community about the Feriji project. We selected 104 representative Zarma speakers, including both native and non-native speakers. The demographics of the survey participants are illustrated in Figures \ref{fig:gender} and \ref{fig:age}. The survey results are summarized in Section \ref{sec: survey}. In addition to the survey responses, the community raised concerns about two key areas: the fluency of the translations and the accessibility of the tool for illiterate Zarma speakers.

\subsection{Survey Responses}\label{sec: survey}

\begin{table*}[t]
\centering
\small
\begin{tabular}{lccc}
\hline
\textbf{Survey Question} & \textbf{Yes} & \textbf{No} & \textbf{Undecided} \\[0.1cm]  
\hline

Does the Feriji project effectively address the linguistic needs of the Zarma community? & 99 & 5 & 0 \\[0.1cm]  
\hline

Do you see Feriji supporting educational initiatives in Zarma-speaking regions? & 98 & 0 & 6 \\[0.1cm]  
\hline

Do you think Feriji will significantly impact preserving the cultural heritage of the Zarma people? & 100 & 4 & 0 \\[0.1cm]  
\hline

Do you foresee any challenges or barriers to the widespread adoption of Feriji within the community? & 45 & 61 & 0 \\[0.1cm]  
\hline

Are you likely to recommend the Feriji project to others within your community? & 102 & 0 & 2 \\[0.1cm]  
\hline
\end{tabular}
\caption{Feriji Community Survey Results}
\label{tab:feriji_survey}
\end{table*}

As shown in Table \ref{tab:feriji_survey}, the survey results indicate strong community support for and optimism about the Feriji project. A significant majority (95\%) believe that Feriji effectively addresses the linguistic needs of the Zarma community. Additionally, 94.2\% of participants believe that Feriji can support educational initiatives in Zarma-speaking regions. Further, 96.2\% of respondents are confident that Feriji will significantly impact preserving the cultural heritage of the Zarma people.
Despite these positive responses, the survey also revealed concerns about the widespread adoption of Feriji. 43.3\% of participants anticipated challenges or barriers to implementation, mainly due to the high illiteracy rate in the region. Nonetheless, 98.1\% of respondents indicated they would recommend Feriji to others in their community.
These findings demonstrate the perceived value of the Feriji project and provide valuable insights for its future development, as emphasized by \citet{nekoto2020participatory}.

\subsection{Translation Fluency Feedback}
Community members acknowledged our efforts to improve translation fluency but noted that the translations were only sometimes fluent. This is a common challenge in MT projects involving low-resource languages, as highlighted by \citet{koehn2017six}. Community members suggested that we focus on expanding and diversifying the training data to enhance the fluency of the translations.

\subsection{Accessibility Concerns}
Another theme in the feedback was the accessibility of the Translator for illiterate members of the Zarma community. This concern aligns with broader challenges of inclusivity for language technology, as discussed by \citet{medhi2011designing}. Community members proposed developing a text-to-speech (TTS) system to address this issue, drawing inspiration from successful implementations in other under-resourced languages \cite{ogayo2022building}.

\section{Areas of Application}
The Feriji Translator has potential applications that can benefit the Zarma-speaking community. This section highlights some key areas where FT can be effectively used.

\subsection{Educational Content Translation}
One primary application of FT is the translation of educational materials. Such materials are often available in French, posing comprehension difficulties for native Zarma speakers. Feriji can make these materials accessible in Zarma, thereby enhancing understanding and learning effectiveness. This is supported by \citet{unesco2016understand}, who found that students receiving instruction in their native language perform significantly better than those receiving instruction in a foreign language.


\subsection{Community Outreach and Public Information}
Public announcements, safety messages, and government communications translated into Zarma can reach a wider audience. This is particularly important in emergencies, where clear and timely communication is crucial. \citet{piller2020linguistic} highlight the importance of language accessibility in public information dissemination in multilingual societies.

\subsection{Cultural Preservation and Promotion}
Feriji can play a major role in preserving and promoting Zarma culture. It can facilitate the translation of literature and historical texts, ensuring their accessibility and preservation for future generations. This is supported by \citet{zhang2022nlp}, who reviewed digital tools in cultural conservation and found that MT technology can be valuable for preserving and promoting endangered languages.

\section{Ethical Considerations}
The development and implementation of MT systems like Feriji raise several ethical considerations that require careful attention.
One primary concern is the potential for cultural insensitivity or misrepresentation, especially when working with languages deeply intertwined with cultural identities, such as Zarma. As highlighted by \citet{hershcovich2022challenges}, MT systems can inadvertently perpetuate stereotypes or misinterpret cultural nuances. This can significantly impact the perception and understanding of a language and its speakers. To mitigate this risk, we engaged closely with native Zarma speakers and cultural experts throughout the development process to ensure that Feriji is culturally sensitive and respectful.
Another critical aspect is data privacy and consent, mainly when sourcing texts from the community or online platforms. \citet{rogers2021just} emphasize that the ethical collection and use of data are imperative for maintaining the community's trust and respecting individual rights. In creating Feriji, we adhered to strict guidelines for data collection, ensuring that all sourced materials were publicly available or used with explicit permission.
Furthermore, as MT technology advances, the risk of language homogenization becomes more pronounced, potentially leading to the erosion of linguistic diversity. \citet{bird2020decolonising} addresses this concern, noting the importance of developing MT systems that support---rather than supplant---the richness of indigenous languages. Feriji aims to enhance the accessibility of Zarma while preserving its unique linguistic characteristics.
Lastly, equitable access to technology is a crucial consideration. \citet{vandijk2020digital} points out that advancements in digital technologies often disproportionately benefit those with higher access to technology, exacerbating the divide. Feriji is designed to bridge this gap, making MT technology accessible to Zarma speakers with limited resources.
We aim to ensure Feriji is used responsibly and ethically, benefiting the Zarma community while respecting their privacy, culture, and language.

\section{Conclusion}
This paper introduced Feriji, the first parallel French-Zarma corpus and glossary designed for machine translation. Feriji significantly contributes to the field by addressing the lack of resources for Zarma, a language spoken by over 5 million people in Niger and neighboring countries.
Feriji will be a valuable resource for researchers and developers working on Zarma MT. We anticipate that Feriji will contribute to the promotion of the Zarma language and make it more accessible to people around the world.

\section{Future Work}
Zarma, like many other African languages, is complex. Accurately representing it in MT systems according to its linguistic rules is a significant challenge. The next phase of the Feriji project will focus on creating a disambiguation tool called Hanse\ny an. This tool will either be based on pattern-based morphemic analysis \cite{Jarad_2015} or trained as an ML model to correct grammar errors.
In addition to developing Hanse\ny an, we will continue to improve FD. We will release new dataset versions with higher-quality and more diverse sentences, moving away from single-topic-centric content---stories centered on a single theme. We believe these improvements will further enhance the value of FD for researchers and developers working on Zarma MT.

\section{Acknowledgement}
We extend our deepest gratitude to the volunteers who participated in our survey and provided feedback, helping us refine and improve the Feriji project.
We also thank the Computer Science department of Ashesi University for providing financial support and cloud resources.
We are grateful to everyone who contributed to the creation of Feriji and supported our efforts to promote and preserve the Zarma language.

\bibliography{anthology,custom}
\bibliographystyle{acl_natbib}

\appendix
\section{Detailed Data Collection Pipeline}
\label{Detailed data collection}

\subsection{Data Sources and Distribution}
The FD comprises data collected from various sources, including religious texts, materials from the Peace Corps, and stories generated using ChatGPT. The distribution of the data sources in the FD is as follows: 70\% of Religious Texts, 20\% Peace Corps Materials, and 10\% ChatGPT Generated Stories. The prompte structure for generating the stories can be found in Subsection \ref{prompt}.

\subsection{ChatGPT Prompt Structure}
\label{prompt}
To ensure culturally appropriate and accurate content, we designed prompts for generating stories with ChatGPT. The prompts included specific details such as the names of characters, the setting, and the scenario.

\begin{tcolorbox}[colback=gray!10!white, colframe=gray!80!black, title=Example Prompt]
\textit{
Crée une nouvelle se déroulant dans un village du Niger. L'histoire doit comprendre trois personnages principaux : Moussa, un jeune garçon, Amina, sa jeune sœur, et Habi, leur cousine. Le cadre est un village africain typique avec des constructions de types traditionel, une place de marché centrale et le fleuve Niger à proximité. L'histoire doit tourner autour de Moussa qui apprend à Amina et Habi à pêcher dans le fleuve. Inclue des dialogues et des descriptions qui reflètent l'environnement culturel et la vie quotidienne d'une communauté.
}
\end{tcolorbox}

\begin{itemize}
    \item \textbf{Names of characters:} Moussa, Amina, Habi
    \item \textbf{Setting:} A village in Niger with traditional buildings, a central market place, and the Niger River nearby
    \item \textbf{Scenario:} Moussa teaching Amina and Habi how to fish in the river
    \item \textbf{Other details:} Includes dialogues and descriptions reflecting the cultural environment and daily life of the community
\end{itemize}

\subsection{Data Cleaning and Initial Automatic Alignment}
The initial collected data---from online sources---contained noise and missing translations, which required a series of cleaning steps to remove the tags---xml tags. We then used custom python scripts to automatically align the French and Zarma sentences. These scripts removed duplicates, and ensured the sentences were properly paired.

\subsection{Human Review Process}
Human reviewers played an important role in verifying the accuracy and cultural appropriateness of the data. The review process---for both online and generated data---included the following steps:

\begin{enumerate}
    \item \textbf{Review of Online Sources:}
    \begin{itemize}
        \item Reviewers cross-checked sentences collected from online sources to ensure proper alignment after the initial automatic alignment.
        \item They read through the aligned sentences, correcting any mistakes and ensuring the translations were accurate and culturally appropriate.
    \end{itemize}
    \item \textbf{Review of ChatGPT Generated Stories:}
    \begin{itemize}
        \item Reviewers initially read the stories in French to ensure they were culturally appropriate and free from bias or offensive content.
        \item They translated the stories into Zarma, maintaining the cultural context and accuracy.
        \item Reviewers then aligned the French and Zarma versions of the stories.
    \end{itemize}
\end{enumerate}

\section{Detailed Human Evaluation Process}
\subsection{Recruitment Process}
For the human evaluation process, we recruited five native Zarma speakers to participate as evaluators. The recruitment was conducted on a volunteer basis, and no monetary compensation was provided to the participants. The evaluators were selected to ensure a diverse representation in terms of age and gender.

\subsection{Training Provided to Evaluators}
To ensure the evaluators were well-prepared for the task, we provided a brief training session before the evaluation began. The training included:

\begin{itemize}
    \item An overview of the evaluation criteria: fluency, comprehension, and readability.
    \item Examples of translations with varying levels of quality to illustrate the rating scale from 1 to 5.
    \item A practice session where evaluators rated a small set of translations and discussed their ratings to align their understanding of the criteria.
\end{itemize}

\subsection{Measures of Inter-Annotator Agreement}
Inter-annotator agreement is necessary for ensuring the reliability of human evaluation. To measure this agreement, we calculated the mean and standard deviation of the scores provided by the evaluators, as shown in Table \ref{table:detailed_evaluation} in the main text. The consistency of the scores across evaluators was analyzed to assess the level of agreement.

The analysis of the mean and standard deviation values indicates that there was a high level of agreement among the evaluators. The low standard deviation values suggest that the ratings were consistent across different evaluators, reinforcing the reliability of the human evaluation process.

\end{document}